# A rigid-body motion reduction technique to improve the convergence of the physics-informed neural networks in solid mechanics

Alireza Sadeghirad[1*], Ali Vaziri[2]

[1]*Department of Civil and Environmental Engineering, Amirkabir University of Technology, Tehran, Iran*

[2]*Independent Researcher, Austin, Texas, USA*

**Abstract**

This work presents a novel approach to enhance the convergence of physics-informed neural networks (PINNs) for solid mechanics problems by utilizing a rigid-body motion reduction (RBMR) technique. Although known to harness the physical laws into machine learning models, PINNs frequently suffer from imbalances in the loss function and rigid-body motion artifacts. There are solutions to such challenges in our RBMR approach: it incorporates a modified collocation loss function allowing for a few arbitrary points to be anchored to eliminate the rigid-body motion of the problem analytically, which ensures physical realism. Validation of the proposed method is performed on numerical experiments conducted on 2D elastostatic problems, including a pure bending cantilever beam as well as a curved cantilever beam under end loading. The results show performance substantially improved in accuracy and convergence against the conventional PINN approach, revealing that digesting specific domain insights inside the training process proved beneficial.

[*] Corresponding author. E-mail: sadeghirad@aut.ac.ir

## 1. Introduction

Neural networks (NNs), modeled after the human brain's biological architecture, have attracted considerable interest for their exceptional ability to detect complex patterns and make predictions from data [1]. These models are built from interconnected nodes, or neurons, organized into layers, where each layer performs a simple mathematical function. Through training, NNs fine-tune their internal parameters to narrow the gap between predicted and actual outcomes. NNs have shown their effectiveness over time in various fields, such as chemistry [2], physics [3], biology [4], computer vision [5], and language processing [6].

Physics-informed neural networks (PINNs) are a class of neural networks designed to incorporate physical laws directly into the learning process for solving partial differential equations (PDEs) [7]. This approach allows the network to learn from both data and the underlying physics, reducing the dependency on large datasets. PINNs offer a robust approach for solving complex nonlinear problems [8], bypassing traditional issues like relying on spatial grid by using automatic differentiation (AD) [9] to obtain partial derivatives. Various frameworks have been developed for PINNs to solve both forward and inverse problems across a range of computational mechanics frameworks [10-14]. Electromagnetic problem [15], elastodynamics [16], and seismic wave modeling [17], are among the prominent areas where PINNs enable robust simulations. Additionally, fields such as fluids mechanics [18], bio-engineering [19], material science [20], fracture mechanics [21], and uncertainty quantification [13], benefit from PINNs, as they offer flexible frameworks to approximate PDE solutions in complex systems, even extending to quantum mechanics [22], high-dimensional [23], and fractional differential equations [24]. A comprehensive review of PINNs, their applications, and extended variants can be found in, e.g., [25-27].

Despite the great success of PINNs in solving complex PDEs, they still face limitations, though remedies continue to be proposed. High computational costs arise from the expensive AD computations required for high-dimensional problems [11]. Furthermore, AD can lead to the gradient instability when computing higher-order derivatives, which in turn may destabilize network training [28, 29]. To mitigate these limitations, several approaches have been proposed. Notable examples include the smoothing kernel PINN (SK-PINN) [30], the gradient-smoothing PINN (GS-PINN) [31], the Physics-Informed Graph Ordinary Differential Equation (PIGODE) model [32], the coupled-automatic–numerical PINN (CAN-PINN) [33], the hybrid finite element and PINN (FEINN) method [34], the first-order PINN (FO-PINN) [29], and the Tanh-B6 KAN-based PINN [35]. Discrete PINNs, such as those using convolutional neural networks (CNNs) have also been investigated to improve the efficiency and accuracy of PINNs [36]. Accuracy issues may also arise when handling stiff PDEs [37], which can be alleviated using Self-Adaptive PINNs [38]. Furthermore, fully connected PINN architectures have limited ability to learn functions with high-frequency components due to spectral bias [39]. Additional limitations are discussed in, e.g., [26, 27, 40, 41].

Application of the PINN methods on solid mechanics problems have been reviewed by some authors [42, 43], with a focus on exploring different loss function formulations. For solid-mechanics problems, the collocation loss function, introduced in [44], is based on the strong form of equations and minimizes the residuals of governing equations and boundary conditions at sampling points. However, this can lead to an imbalance in scales between different residual terms, which can hinder model performance. To address this issue, various strategies have been proposed, including adjustable training weights [45], annealing learning rate [37], and dynamic noise parameters [46]. A notable advancement in this area was the introduction of neural tangent kernel (NTK) theory to PINNs [47, 48], resulting in an NTK-based adaptive learning scheme that dynamically balances the loss terms. This approach has shown promise in handling more complex multi-scale [49] and long-term evolution problems [50]. An effective alternative is the energy-based loss function, which is rooted in the weak-form formulation and directly minimizes potential energy. This method naturally aligns the scales of loss terms, as they all represent energy quantities, and has been applied to many problems, including Kirchhoff plates [51] and crack path prediction [52]. Despite these advantages, the energy-based loss function can struggle with accurately capturing fine stress fields, as it does not embed equilibrium conditions and boundary requirements explicitly. Furthermore, to enhance model fidelity, other loss formulations have been proposed, such as integrating the energy-based loss function with the collocation loss function [53], combining the collocation loss function with work balance loss [54], and using the least squares weighted residual (LSWR) method [55] to address scale-difference issues.

Our research contributes to the established concept of incorporating prior knowledge into PINNs by proposing a modified loss function that enhances the network's ability to capture physical solutions. Specifically, we focus on modifying the collocation loss function for solid mechanics to eliminate rigid body motion from the solution. By anchoring the solution using a few arbitrary points within the domain, this approach provides the network with insights into the underlying physical behavior. The use of arbitrary anchor points makes the method applicable to a wider range of problems and geometries with different boundary conditions. To validate the effectiveness of our modified loss function, we conducted a series of numerical experiments on 2D elastostatic problems, including a pure bending cantilever beam and a curved cantilever beam under end loading. These problems were chosen due to their well-established analytical solutions, allowing for a robust comparison between our modified PINN and standard PINN approaches. For comparison with the standard PINN, we apply NTK-based training to both the modified loss function and the standard PINN. Additionally, we use optimal parameters for each method, obtained through comprehensive hyperparameter tuning. The comparison of relative errors across multiple runs demonstrated that our approach yielded more accurate predictions of displacements and stresses, underscoring the importance of embedding domain-specific insights within the loss function.

The outline of this paper is as follows: Section (2) provides a brief review of the standard PINN method. The proposed RBMR technique is discussed in Section (3). Finally, some numerical examples are presented in Section (4), followed by some conclusions made in Section (5).

## 2. Review of the standard PINN method

### 2.1 *Physics-informed neural networks*

In this section, we summarize the foundational concepts underlying standard PINNs [7]. These networks are designed to solve differential equations that are generally expressed in the following form:

$$\mathcal{N}[u](\boldsymbol{x}) = f(\boldsymbol{x}), \qquad \boldsymbol{x} \in \Omega$$
$$\mathcal{B}[u](\boldsymbol{x}) = g(\boldsymbol{x}), \qquad \boldsymbol{x} \in \partial\Omega \tag{1}$$

Here, $\mathcal{N}$ represents a differential operator, $u(\boldsymbol{x})$ is the solution function, $f(\boldsymbol{x})$ is a source term, $\Omega$ denotes the spatial domain for $\boldsymbol{x} = (x_1, \ldots, x_d)$, $\mathcal{B}$ encodes boundary conditions operator (e.g. Dirichlet, Neumann, Robin, etc.), and $g(\boldsymbol{x})$ is the boundary function. For initial-boundary value problems, time can be treated as an additional coordinate alongside $\boldsymbol{x}$ , with $\Omega$ representing the combined spatio-temporal domain. A feedforward neural network $\mathcal{F}$ is employed to approximate the solution function:

$$u(\boldsymbol{x}) = \mathcal{F}(\boldsymbol{x}, \boldsymbol{\theta}) \tag{2}$$

where $\boldsymbol{\theta}$ denotes the set of all trainable parameters in the network, enabling residual to be defined as:

$$r(\boldsymbol{x}, \boldsymbol{\theta}) = \mathcal{N}[u](\boldsymbol{x}, \boldsymbol{\theta}) - f(\boldsymbol{x}) \tag{3}$$

The network can be determined by minimizing a loss function, i.e.

$$\boldsymbol{\theta} = \arg\min_{\boldsymbol{\theta}} \mathcal{L}(\boldsymbol{\theta}) \tag{4}$$

where

$$\mathcal{L}(\boldsymbol{\theta}) = \mathcal{L}_r(\boldsymbol{\theta}) + \mathcal{L}_b(\boldsymbol{\theta}) \tag{5}$$

in which $\mathcal{L}_r(\boldsymbol{\theta})$ and $\mathcal{L}_b(\boldsymbol{\theta})$ are the loss terms from the PDE and boundary conditions, respectively.

Using the collocation loss function, the residual terms are minimized to zero at sampling points, with the mean square error (MSE) serving as the basis:

$$\mathcal{L}_r(\boldsymbol{\theta}) = \frac{1}{2}\sum_{i=1}^{N^r} |\mathcal{N}[u](\boldsymbol{x}_i, \boldsymbol{\theta}) - f(\boldsymbol{x}_i)|^2$$
$$\mathcal{L}_b(\boldsymbol{\theta}) = \frac{1}{2}\sum_{i=1}^{N^b} |\mathcal{N}\mathcal{B}[u](\boldsymbol{x}_i, \boldsymbol{\theta}) - g(\boldsymbol{x}_i)|^2 \tag{6}$$

where $N^r$ and $N^b$ are the number of all sampling points and the number of boundary sampling points, respectively. Boundary conditions are imposed softly by treating them as penalty terms within the loss function. This approach ensures that the BC constraints contribute to the optimization process through additional loss components defined on the boundary sampling points [7].

### 2.2 *Neural tangent kernel*

Given different terms in the loss function, the neural tangent kernel (NTK) [47, 48] can be utilized to guide the weighting of these terms, enhancing performance. This is accomplished by balancing the components of the PINN loss function in Eq. (4) to ensure that all constraints are satisfied uniformly. For this optimization problem, selecting the gradient descent (GD) method with a sufficiently small learning rate and other assumptions discussed in [47], transforms the optimization process into a gradient flow system,

$$\partial \boldsymbol{\theta}^T / \partial t = -\nabla_\theta \mathcal{L}(\boldsymbol{\theta}) \tag{7}$$

Now, training a PINN through GD can be rewritten as the following kernel gradient descent:

$$\begin{bmatrix} \mathcal{N}_t[u](\boldsymbol{x}_r, \boldsymbol{\theta}) \\ \mathcal{B}_t[u](\boldsymbol{x}_b, \boldsymbol{\theta}) \end{bmatrix} = -\mathbf{K} \begin{bmatrix} \mathcal{N}[u](\boldsymbol{x}_r, \boldsymbol{\theta}) - f(\boldsymbol{x}_r) \\ \mathcal{B}[u](\boldsymbol{x}_b, \boldsymbol{\theta}) - g(\boldsymbol{x}_b) \end{bmatrix} \tag{8}$$

where $t$ indicates the training evolution, and $\mathbf{K} = [\mathbf{K}_{rr}, \mathbf{K}_{rb}; \boldsymbol{K}_{rb}^T, \mathbf{K}_{bb}]$ is the NTK of the PINN, with each entry defined as:

$$\begin{aligned} K_{rr}^{ij} &= ((\nabla_\theta \mathcal{N}[u](\boldsymbol{x}_r^i, \boldsymbol{\theta}))^T \nabla_\theta \mathcal{N}[u](\boldsymbol{x}_r^j, \boldsymbol{\theta})) \\ K_{rb}^{ij} &= ((\nabla_\theta \mathcal{N}[u](\boldsymbol{x}_r^i, \boldsymbol{\theta}))^T \nabla_\theta \mathcal{B}[u](\boldsymbol{x}_b^j, \boldsymbol{\theta})) \\ K_{bb}^{ij} &= ((\nabla_\theta \mathcal{B}[u](\boldsymbol{x}_b^i, \boldsymbol{\theta}))^T \nabla_\theta \mathcal{B}[u](\boldsymbol{x}_b^j, \boldsymbol{\theta})) \end{aligned} \tag{9}$$

The spectral decomposition of the NTK, $\mathbf{K} = \mathbf{Q}\boldsymbol{\Lambda}\mathbf{Q}^T$, can be applied in the adaptive optimization process to balance the terms in the loss function and ensure uniform satisfaction of all constraints, where $\mathbf{Q}$ is an orthogonal matrix and $\boldsymbol{\Lambda}$ is a diagonal matrix of eigenvalues. We apply this approach to both the standard PINN and the method discussed in Sec. 3. Details of the derivation and further theoretical discussions can be found in [47, 48].

### 2.3 *Model problem*

In this section, the model problem of two-dimensional linear elasticity is considered, which is governed by the following system of equations:

$$\begin{aligned} \boldsymbol{\nabla} \cdot \boldsymbol{\sigma} + \boldsymbol{b} = \boldsymbol{0}, &\qquad \boldsymbol{x} \in \Omega \\ \boldsymbol{\sigma} \cdot \boldsymbol{n} = \bar{\boldsymbol{t}}, &\qquad \boldsymbol{x} \in \Gamma_N \\ \boldsymbol{u} = \bar{\boldsymbol{u}}, &\qquad \boldsymbol{x} \in \Gamma_D \end{aligned} \tag{10}$$

where $\boldsymbol{\nabla}$ represents two-dimensional nabla operator, $\boldsymbol{\sigma}$ is the stress tensor, $\boldsymbol{b} = [b_x, b_y]^T$ is the body force, and $\boldsymbol{u} = [u_x, u_y]^T$ is the vector of infinitesimal in-plane displacements. Dirichlet and Neumann boundary conditions are denoted by $\Gamma_D$ and $\Gamma_N$, respectively, where $\Gamma_D$ corresponds to the prescribed displacement $\bar{\boldsymbol{u}}$ and $\Gamma_N$ corresponds to the prescribed traction $\bar{\boldsymbol{t}}$, with the normal vector $\boldsymbol{n}$. The strain tensor $\boldsymbol{\varepsilon}$ can be calculated from the displacement field $\boldsymbol{u}$ as

$$\boldsymbol{\varepsilon} = \frac{1}{2}[(\boldsymbol{\nabla u}) + (\boldsymbol{\nabla u})^{\mathrm{T}}] \tag{11}$$

In isotropic linear elasticity under plane strain conditions, the stress vector is related to the strain vector through the linear relationship:

$$\begin{bmatrix} \sigma_{xx} \\ \sigma_{yy} \\ \sigma_{xy} \end{bmatrix} = \begin{bmatrix} \lambda + 2\mu & \lambda & 0 \\ \lambda & \lambda + 2\mu & 0 \\ 0 & 0 & \mu \end{bmatrix} \begin{bmatrix} \varepsilon_{xx} \\ \varepsilon_{yy} \\ 2\varepsilon_{xy} \end{bmatrix} \tag{12}$$

in which $\lambda$ and μ are the Lamé constants, which are related to the Young's modulus $E$ and Poisson's ratio $\nu$ as

$$\lambda = \frac{\nu E}{(1+\nu)(1-2\nu)}$$
$$\mu = \frac{E}{2(1+\nu)} \tag{13}$$

Substituting Eqs. (11) and (12) into the equilibrium equation (Eq. (10)) results in the Navier-Lamé PDEs under the plane strain conditions:

$$(\lambda+2\mu)\frac{\partial^2 u_x}{\partial x^2} + \mu\frac{\partial^2 u_x}{\partial y^2} + (\lambda+\mu)\frac{\partial^2 u_y}{\partial x \partial y} + b_x = 0$$
$$(\lambda+2\mu)\frac{\partial^2 u_y}{\partial y^2} + \mu\frac{\partial^2 u_y}{\partial x^2} + (\lambda+\mu)\frac{\partial^2 u_x}{\partial x \partial y} + b_y = 0 \tag{14}$$

Note that for the problem at hand, with Dirichlet and Neumann boundary conditions, we split the contribution to the boundary terms of the loss function in Eq. (5). Therefore, the total loss can be expressed as:

$$\mathcal{L}(\boldsymbol{\theta}) = \mathcal{L}_r(\boldsymbol{\theta}) + \mathcal{L}_{\mathrm{NBC}}(\boldsymbol{\theta}) + \mathcal{L}_{\mathrm{DBC}}(\boldsymbol{\theta}) \tag{15}$$

where

$$\mathcal{L}_{\mathrm{NBC}}(\boldsymbol{\theta}) = \frac{1}{2}\sum_{i=1}^{N^n} |\boldsymbol{\sigma}(\boldsymbol{x}_i, \boldsymbol{\theta}) \cdot \boldsymbol{n} - \bar{\boldsymbol{t}}_i|^2$$
$$\mathcal{L}_{\mathrm{DBC}}(\boldsymbol{\theta}) = \frac{1}{2}\sum_{i=1}^{N^d} |\boldsymbol{u}(\boldsymbol{x}_i, \boldsymbol{\theta}) - \bar{\boldsymbol{u}}_i|^2 \tag{16}$$

where $N^n$ and $N^d$ are the number of boundary sampling points corresponding to Neumann and Dirichlet boundary conditions, respectively.

## 3. The rigid-body motion reduction (RBMR) technique

Two types of motion can be recognized in continuum-mechanics, including strain deformation and rigid-body motion. In rigid-body motion, the distance between points remains the same. This motion does not contribute to the strain field, and thus also does not affect the stress field. In many solid-mechanics problems, such as static equilibrium, structural analysis, vibration analysis, and some dynamic problems, inhibiting rigid-body motion is essential to prevent unwanted movements. This is done by applying boundary conditions or constraints that restrict the body's ability to freely translate or rotate. In the standard PINN formulation, these constraints are enforced as part of the Dirichlet boundary conditions during the training process. In this paper, we propose to reformulate the third term in the loss function, $\mathcal{L}_{\mathrm{DBC}}(\boldsymbol{\theta})$ in Eq. (16), by analytically enforcing the suppression conditions of rigid-body motion conditions on the neural network solutions. In this approach, the neural network is not required to converge to the solution without rigid-body motion. Instead, any solution, including arbitrary rigid-body motion, would represent the exact final solution. This modification improves the convergence of the PINN formulation, and the numerical simulations show that this technique yields more accurate results than the standard formulation.

In two-dimensional problems with the assumption of small deformations, the general form of rigid-body displacement can be written as [56]

$$\begin{aligned} u_x^{\text{rigid}} &= u_x^0 - \omega_z^0 y \\ u_y^{\text{rigid}} &= u_y^0 + \omega_z^0 x \end{aligned} \tag{17}$$

in which, $\omega_z^0$ is a constant rotation about the *z*-axis, perpendicular to the plane, and $u_x^0$ and $u_y^0$ are arbitrary constant translations in the *x*- and *y*-directions, respectively.

To define the rigid-body motion, three known displacement components are required, all of which must be neither parallel nor concurrent. Consider a problem domain $\Omega$, shown in Fig. 100, including the Dirichlet and Neumann boundaries. Three known non-concurrent and non-parallel displacement components, $\bar{u}_{n_1}^{\mathrm{A}}$, $\bar{u}_{n_2}^{\mathrm{B}}$, and $\bar{u}_{n_3}^{\mathrm{C}}$, can be chosen arbitrarily as shown in this figure. To find $u_x^0$, $u_y^0$, and $\omega_z^0$, the following system of linear equations can be solved:

$$\begin{aligned} n_1^x u_x^0 + n_1^y u_y^0 - \left(n_1^x y_A - n_1^y x_A\right)\omega_z^0 &= \bar{u}_{n_1}^{\mathrm{A}} \\ n_2^x u_x^0 + n_2^y u_y^0 - \left(n_2^x y_B - n_2^y x_B\right)\omega_z^0 &= \bar{u}_{n_2}^{\mathrm{B}} \\ n_3^x u_x^0 + n_3^y u_y^0 - \left(n_3^x y_C - n_3^y x_C\right)\omega_z^0 &= \bar{u}_{n_3}^{\mathrm{C}} \end{aligned} \tag{18}$$

where $\bar{u}_{n_1}^{\mathrm{A}}$, $\bar{u}_{n_2}^{\mathrm{B}}$, and $\bar{u}_{n_3}^{\mathrm{C}}$ are the prescribed displacement of node A in the direction of $\vec{\boldsymbol{n}}_1$, the prescribed displacement of node B in the direction of $\vec{\boldsymbol{n}}_2$, and the prescribed displacement of node C in the direction of $\vec{\boldsymbol{n}}_3$, respectively. Noting that for the non-concurrent and non-parallel displacement components, the determinant of the coefficient matrix is always non-zero, the system of linear equations (18) always has a unique solution,

$$\Delta = \begin{vmatrix} n_1^x & n_1^y & -\left(n_1^x y_A - n_1^y x_A\right) \\ n_2^x & n_2^y & -\left(n_2^x y_B - n_2^y x_B\right) \\ n_3^x & n_3^y & -\left(n_3^x y_C - n_3^y x_C\right) \end{vmatrix} \neq 0 \tag{19}$$

Actually, the condition $\Delta = 0$ is the necessary condition for the three displacement components to be parallel or concurrent. In this case, if $\frac{n_1^x}{n_1^y} = \frac{n_2^x}{n_3^y} = \frac{n_3^x}{n_3^y}$, the three displacement components are parallel, otherwise they are concurrent.

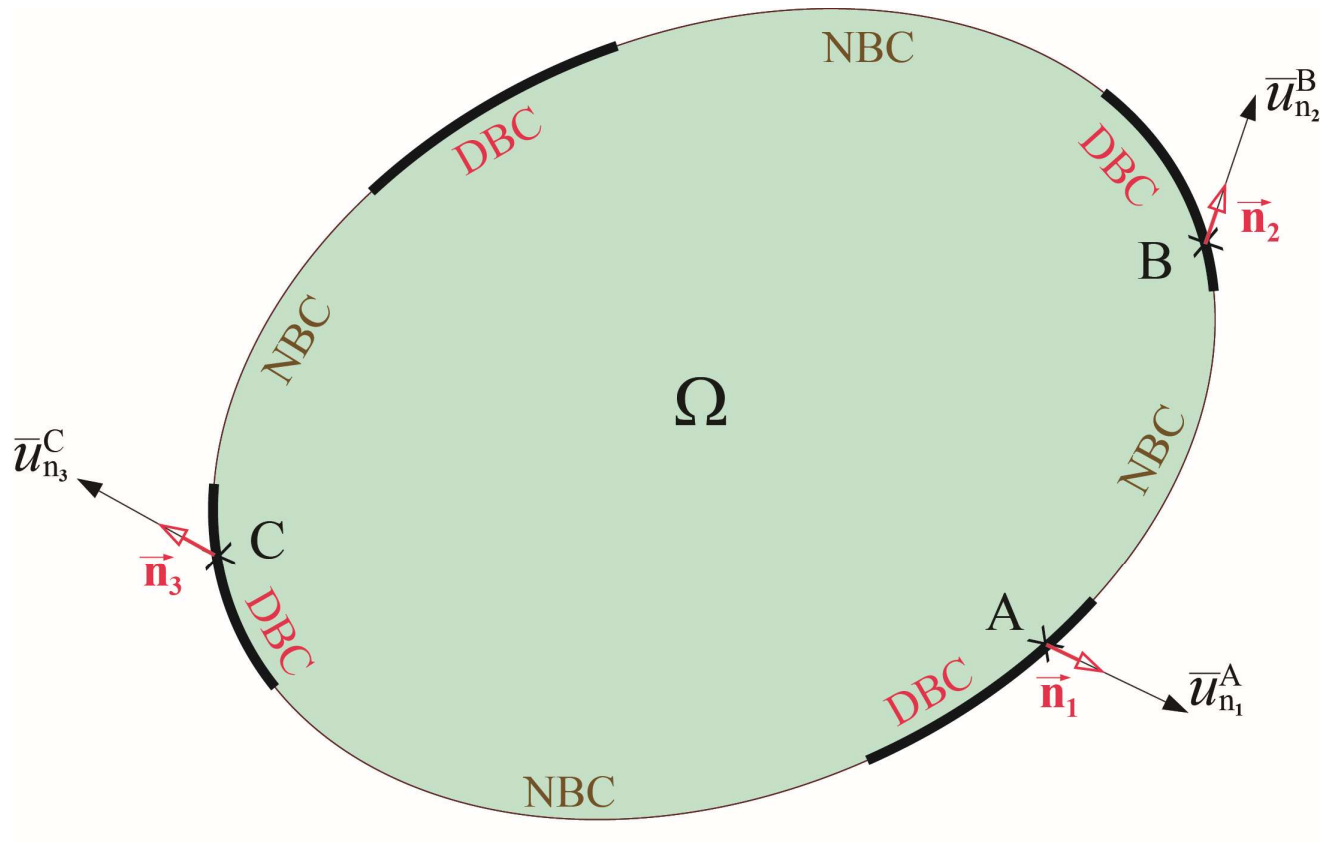

**Fig. 1**. The problem domain Ω with Dirichlet and Neumann boundaries.

Based on the RBMR technique, in each iteration during the training process, three displacement components ${}^{*}u_{n_1}^{\mathrm{A}}$, ${}^{*}u_{n_2}^{\mathrm{B}}$, and ${}^{*}u_{n_3}^{\mathrm{C}}$ are computed from the neural network. The rigid body motion associated with them can be defined similar to Eq. (17) as

$$\begin{aligned} {}^{*}u_x^{\mathrm{rigid}} &= u_x^* - \omega_z^* y \\ {}^{*}u_y^{\mathrm{rigid}} &= u_y^* + \omega_z^* x \end{aligned} \tag{20}$$

where $u_x^*$, $u_y^*$, and $\omega_z^*$ can be calculated, similar to Eq. (18), by solving the following system of linear equations

$$\begin{aligned} n_1^x u_x^* + n_1^y u_y^* - \left(n_1^x y_A - n_1^y x_A\right)\omega_z^* &= {}^{\mathrm{A}}u_{n_1}^{\mathrm{NN}} \\ n_2^x u_x^* + n_2^y u_y^* - \left(n_2^x y_B - n_2^y x_B\right)\omega_z^* &= {}^{\mathrm{B}}u_{n_1}^{\mathrm{NN}} \\ n_3^x u_x^* + n_3^y u_y^* - \left(n_3^x y_C - n_3^y x_C\right)\omega_z^* &= {}^{\mathrm{B}}u_{n_1}^{\mathrm{NN}} \end{aligned} \tag{21}$$

where ${}^{\mathrm{A}}u_{n_1}^{\mathrm{NN}}$, ${}^{\mathrm{B}}u_{n_1}^{\mathrm{NN}}$ and ${}^{\mathrm{C}}u_{n_1}^{\mathrm{NN}}$ are the displacement of node A in the direction of $\vec{\boldsymbol{n}}_1$, the displacement of node B in the direction of $\vec{\boldsymbol{n}}_2$, and the displacement of node C in the direction of $\vec{\boldsymbol{n}}_3$, respectively, computed from the neural network.

Then, the extra rigid body motion is obtained by subtracting Eq. (20) from Eq. (17)

$$\begin{aligned} u_x^{\mathrm{extra}} &= {}^{*}u_x^{\mathrm{rigid}} - u_x^{\mathrm{rigid}} = \Delta u_x^* - \Delta\omega_z^* y \\ u_y^{\mathrm{extra}} &= {}^{*}u_y^{\mathrm{rigid}} - u_y^{\mathrm{rigid}} = \Delta u_y^* + \Delta\omega_z^* x \end{aligned} \tag{22}$$

in which

$$\begin{aligned} \Delta u_x^* &= u_x^* - u_x^0 \\ \Delta u_y^* &= u_y^* - u_y^0 \\ \Delta\omega_z^* &= \omega_z^* - \omega_z^0 \end{aligned} \tag{23}$$

In each iteration during the learning process, this extra rigid body motion is subtracted from the displacements predicted by the neural network in constructing the third term of the loss function as

$$\mathcal{L}_{\mathrm{DBC}}^{\mathrm{RBMR}}(\boldsymbol{\theta}) = \frac{1}{2}\sum_{i=1}^{N^{\mathrm{d}}} |\tilde{\boldsymbol{u}}(\boldsymbol{x}_i, \boldsymbol{\theta}) - \bar{\boldsymbol{u}}_i|^2 \tag{24}$$

where

$$\begin{aligned} \tilde{u}_x &= u_x^{NN} - (\Delta u_x^* - \Delta\omega_z^* y) \\ \tilde{u}_y &= u_y^{NN} - \left(\Delta u_y^* + \Delta\omega_z^* x\right) \end{aligned} \tag{25}$$

in which $u_x^{NN} = u_x\,(\boldsymbol{x}_i, \boldsymbol{\theta})$ and $u_y^{NN} = u_y\,(\boldsymbol{x}_i, \boldsymbol{\theta})$ are the displacement components, computed from the neural network.

The first and second terms in the loss function (Eq. (15)) are not altered since the rigid-body motion does not contribute to the strain and stress fields obtained from the neural network solution. Fig. 2 summarizes the building blocks of the PINN method with the proposed RBMR technique.

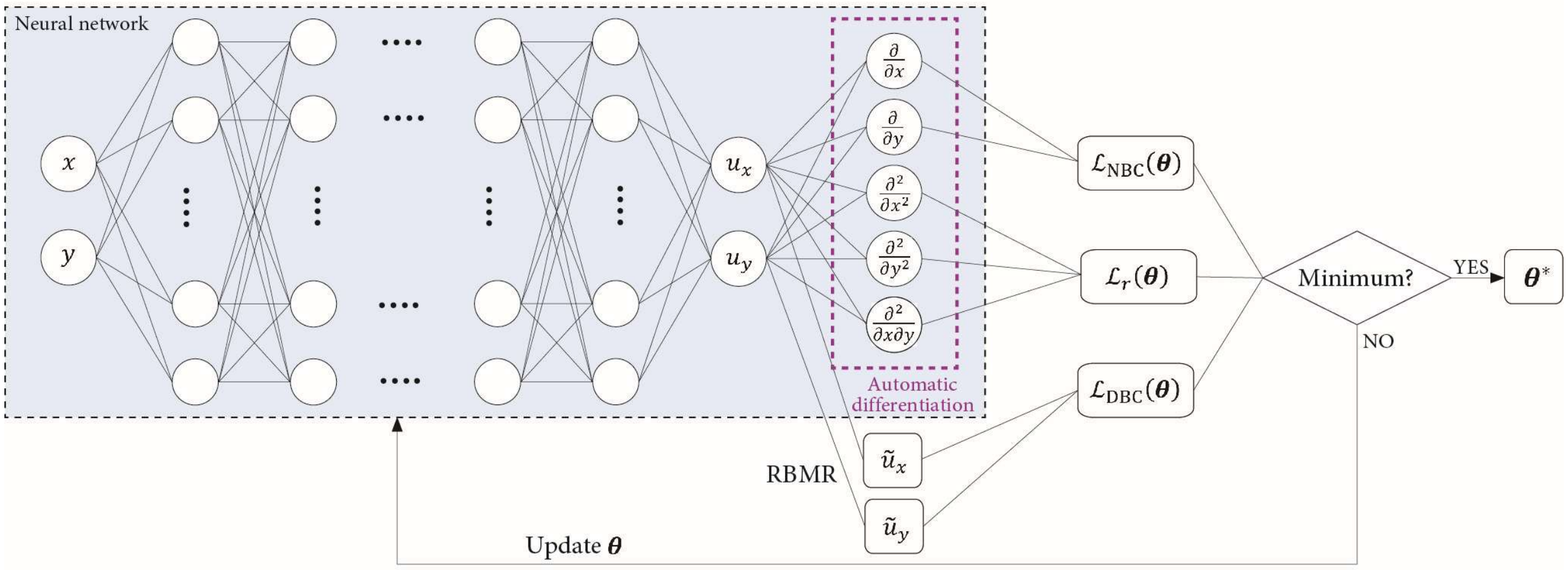

**Fig. 2**. The structure of the PINN method with the proposed RBMR technique.

## 4. Numerical examples

This section presents some numerical examples to study the effectiveness of the proposed RBMR idea in improving the convergence behavior of the PINN. To implement the standard PINN and the proposed method, python codes are built on Tensorflow 2.17 [57]. The source codes, associated with the numerical examples in this section, are available at https://github.com/Sadeghirad/PySolidSimPINN_public. Two feed-forward neural networks (FFNNs) are used to predict two components of the displacement field in the two-dimensional examples. The Swish activation function [58] is used in the FFNNs. This activation function has recently received more attention in the PINN community [25]. To train the neural networks, the Adam optimization scheme [59] is used in this work, which is claimed to be superior to other choices [44] in the PINN simulations. The FFNN inner parameters, i.e. weights and biases, are initialized according to the He initialization scheme [60], available as a Keras kernel initializer in Tensorflow. In this section, each simulation is run 50 times based on different random initializations and the obtained mean error values and their standard deviations are reported. To investigate the convergence of the numerical simulations, the following displacement and energy error norms are adopted:

$$E_u = \left[\frac{\sum_{p=1}^{N^{\mathrm{t}}}\left(\boldsymbol{u}_p^{\mathrm{num}} - \boldsymbol{u}_p^{\mathrm{ex}}\right)\cdot\left(\boldsymbol{u}_p^{\mathrm{num}} - \boldsymbol{u}_p^{\mathrm{ex}}\right) V_p}{\sum_{p=1}^{N^{\mathrm{t}}} \boldsymbol{u}_p^{\mathrm{ex}}\cdot\boldsymbol{u}_p^{\mathrm{ex}}\, V_p}\right]^{\frac{1}{2}} \tag{26}$$

$$E_e = \left[\frac{\sum_{p=1}^{N^{\mathrm{t}}}\left(\boldsymbol{\sigma}_p^{\mathrm{num}} - \boldsymbol{\sigma}_p^{\mathrm{ex}}\right):\left(\boldsymbol{\varepsilon}_p^{\mathrm{num}} - \boldsymbol{\varepsilon}_p^{\mathrm{ex}}\right) V_p}{\sum_{p=1}^{N^{\mathrm{t}}} \boldsymbol{\sigma}_p^{\mathrm{ex}}:\boldsymbol{\varepsilon}_p^{\mathrm{ex}}\, V_p}\right]^{\frac{1}{2}} \tag{27}$$

in which $N^{\mathrm{t}}$ is the total number of sampling points, $V_p$ is the volume of sampling point $p$, and $\boldsymbol{u}_p$, $\boldsymbol{\sigma}_p$, and $\boldsymbol{\varepsilon}_p$ are, respectively, the displacement, stress, and strain fields evaluated at sampling point $p$.

*4.1. Pure bending beam*

In this example, pure bending of a beam is considered. The problem domain is a rectangle with the dimensions of $l \times 2h$ as shown in Fig. 3. Plane strain conditions and unit thickness are assumed here. The analytical stress field for this problem is given as [56]

$$\sigma_x = -\frac{3M}{2h^3}y \qquad \sigma_y = \tau_{xy} = 0 \tag{28}$$

where $M$ is the end moment. The associated displacement field is calculated as

$$u_x = -\frac{3M}{2Eh^3}xy \qquad u_y = \frac{3M}{4Eh^3}(x^2 + \nu y^2) \tag{29}$$

in which $E$ and $\nu$ are the Young's modulus and Poisson’s ratio, respectively.

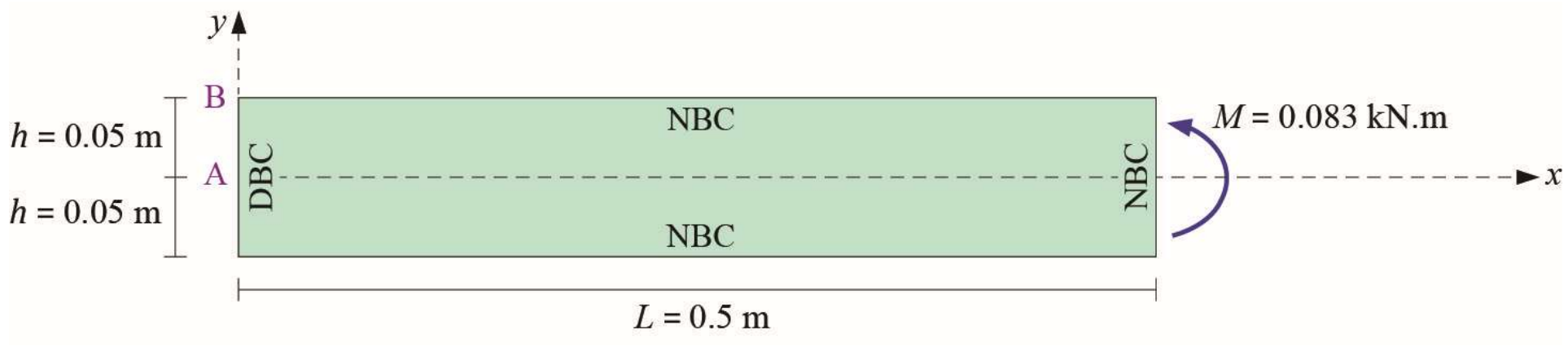

**Fig. 3**. The problem domain in the pure bending example.

In the numerical simulations, Dirichlet boundary conditions are imposed on the left edge of the beam, while Neumann boundary conditions are specified on the remaining boundaries as illustrated in Fig. 1. To implement the RBMR, three known displacement components are chosen as $u_x$ at point $A$, $u_y$ at point $A$, and $u_x$ at point $B$ as shown in this figure. The boundary conditions are defined based on the analytical displacement and stress values from Eqs. (28) and (29). As shown in Fig. 4, two sets of sampling points are used in the numerical simulations, including (1) 96 internal sampling and 60 boundary sampling points, denoted SP1, and (2) 441 internal sampling and 120 boundary sampling points, denoted SP2. Also, two FFNNs with three hidden layers are used in different simulations: (1) FFNN1, which contains 2, 4, 16, 4, and 1 neurons in the input, first hidden, second hidden, third hidden, and output layers, respectively, and (2) FFNN2, which contains 2, 8, 32, 8, and 1 neurons in the input, first hidden, second hidden, third hidden, and output layers, respectively. The Adam optimizer is used with an initial learning rate of 0.001 and an exponential decay with a decay rate of 0.98 and a decay step of 200. The Young's modulus and Poisson's ratio are assumed to be $E = 1000\ \mathrm{kN/m^2}$ and $\nu = 0.3$, respectively.

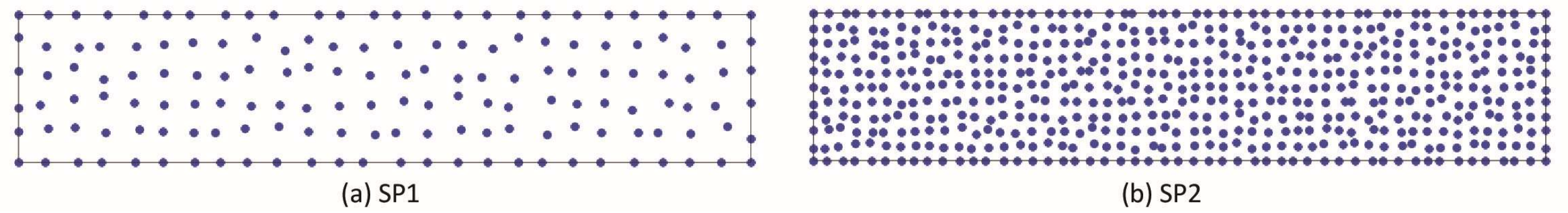

**Fig. 4**. Two sets of sampling points used in the pure bending beam example: (1) 96 internal sampling and 60 boundary sampling points, denoted SP1, and (2) 441 internal sampling and 120 boundary sampling points, denoted SP2.

Each simulation is run 50 times based on different random initializations and the obtained mean error values and their standard deviations are presented in Table 1. Furthermore, Fig. 5 shows the histograms of the displacement and energy error norms for different simulations. It can be seen that the proposed method results in smaller mean errors and standard deviations compared to those obtained from the standard PINN.

Table 1. The mean and standard deviation of the displacement and energy error norms for the pure bending beam example, obtained from the standard PINN and the PINN with RBMR methods

| Neural network | Sampling points | Displacement error norm | | | | Energy error norm | | | |
|---|---|---|---|---|---|---|---|---|---|
| | | standard PINN | | PINN with RBMR | | standard PINN | | PINN with RBMR | |
| | | Mean | Std | Mean | Std | Mean | Std | Mean | Std |
| FFNN1 | SP1 | 0.256 | 0.152 | 0.034 | 0.041 | 0.199 | 0.114 | 0.037 | 0.031 |
| FFNN1 | SP2 | 0.281 | 0.193 | 0.065 | 0.064 | 0.223 | 0.132 | 0.067 | 0.053 |
| FFNN2 | SP1 | 0.209 | 0.109 | 0.013 | 0.013 | 0.166 | 0.086 | 0.016 | 0.010 |
| FFNN2 | SP2 | 0.218 | 0.146 | 0.024 | 0.029 | 0.174 | 0.115 | 0.025 | 0.022 |

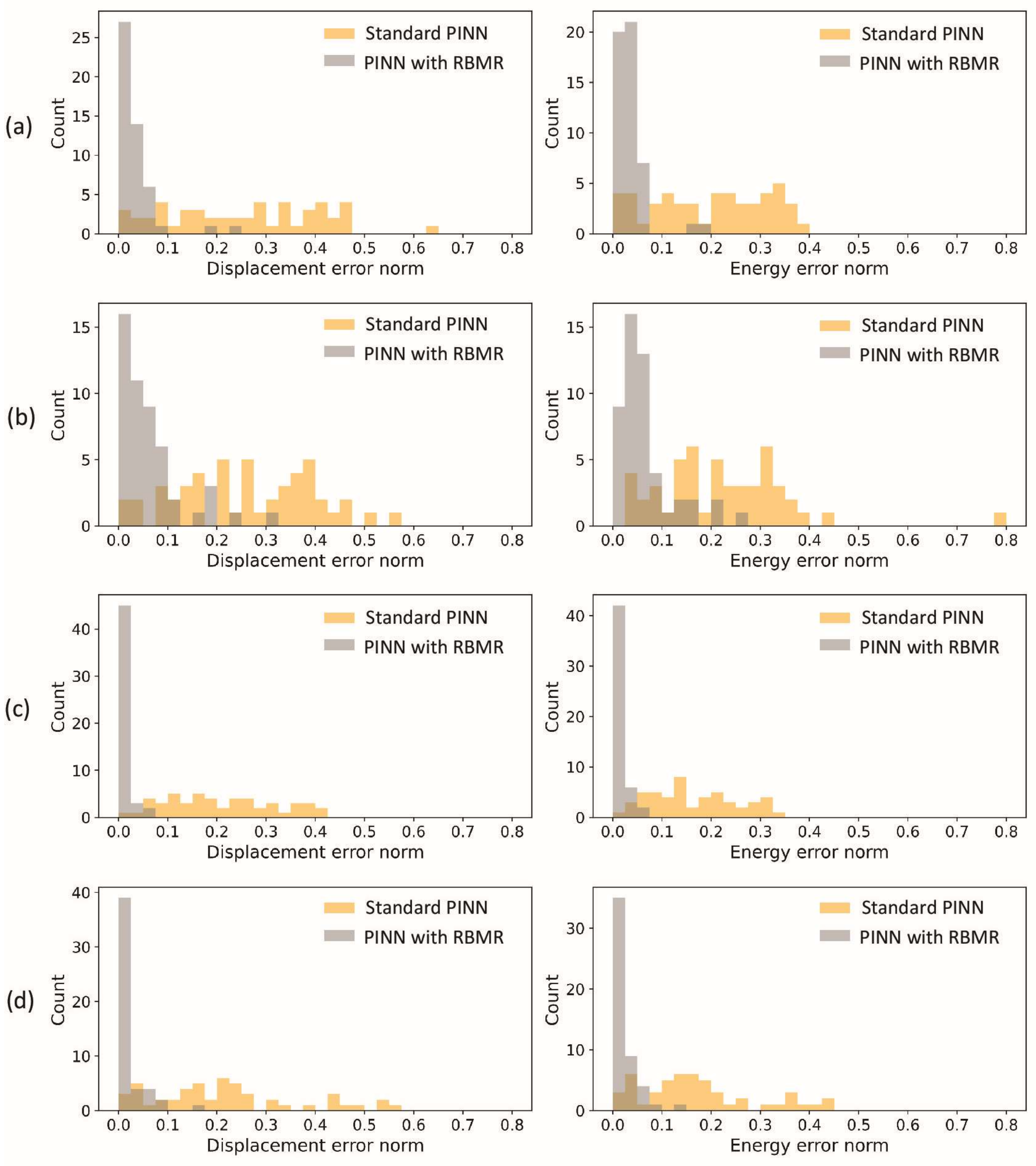

**Fig. 5**. The histograms of the displacement and energy error norms for different simulations of the pure bending beam example with (a) neural network FFNN1 and sampling points SP1, (b) neural network FFNN1 and sampling points SP2, (c) neural network FFNN2 and sampling points SP1, and (d) neural network FFNN2 and sampling points SP2.

The convergence behavior of the simulations is investigated by plotting the mean loss curves in Fig. 6. As shown in this figure the standard PINN and the PINN with RBMR have similar convergence behavior.

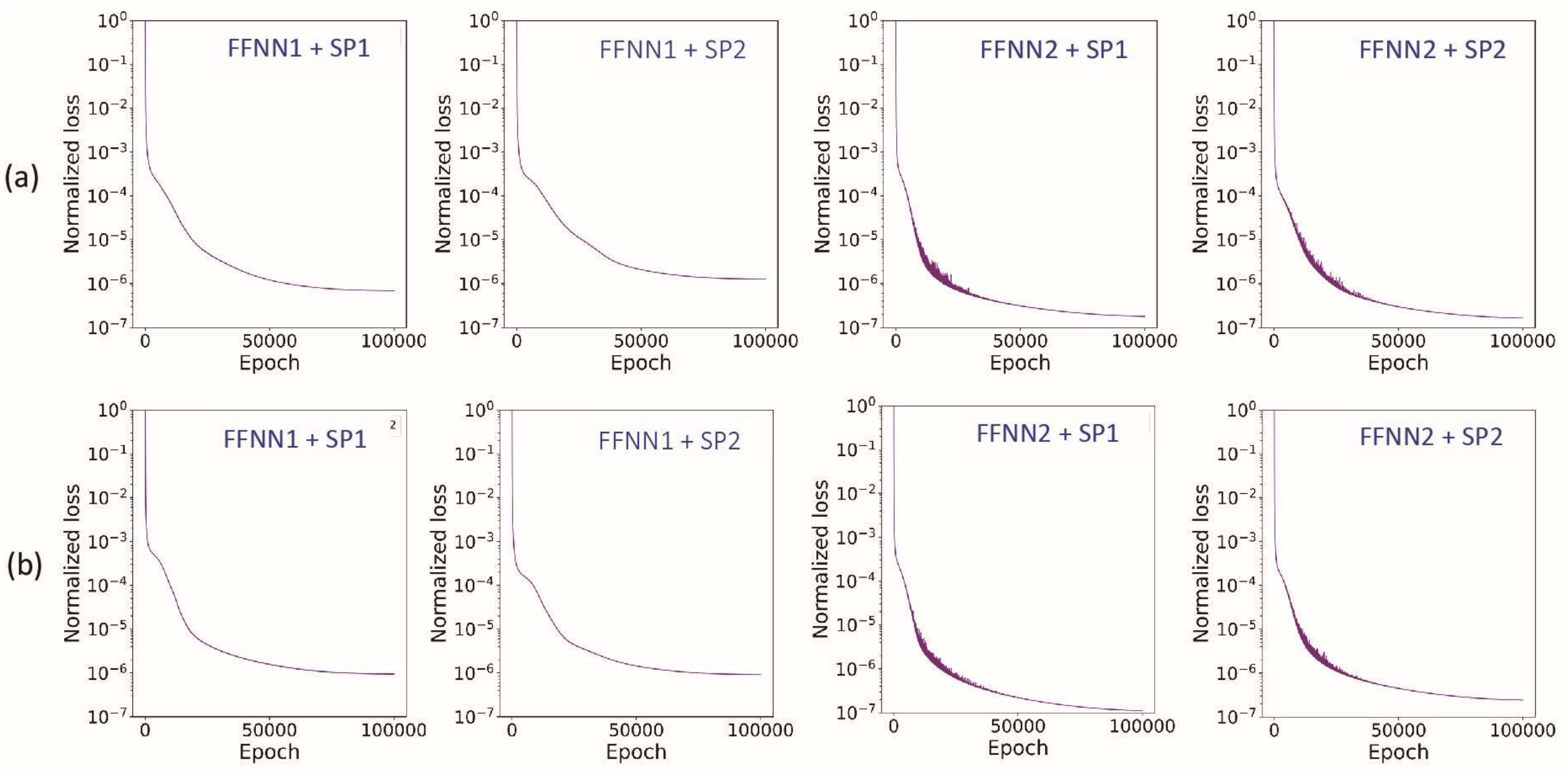

**Fig. 6**. The loss curves for different simulations of the pure bending beam example with (a) standard PINN and (b) PINN with RBMR.

Some snapshots of the simulations with neural network FFNN2 and sampling points SP1 are presented in Fig. 7. These simulations were run by assuming initialization with a seed value of 42. In this case, the obtained displacement norms from the standard PINN and the PINN with RBMR simulations are 0.196 and 0.013 respectively. Similarly, the obtained energy norms from the standard PINN and the PINN with RBMR simulations are 0.153 and 0.020, respectively, which are close to the corresponding mean errors presented in Table 1. As shown in Fig. 7, the results obtained from the PINN with RBMR simulation are in better agreement with the analytical solutions compared to those obtained from the standard PINN simulation.

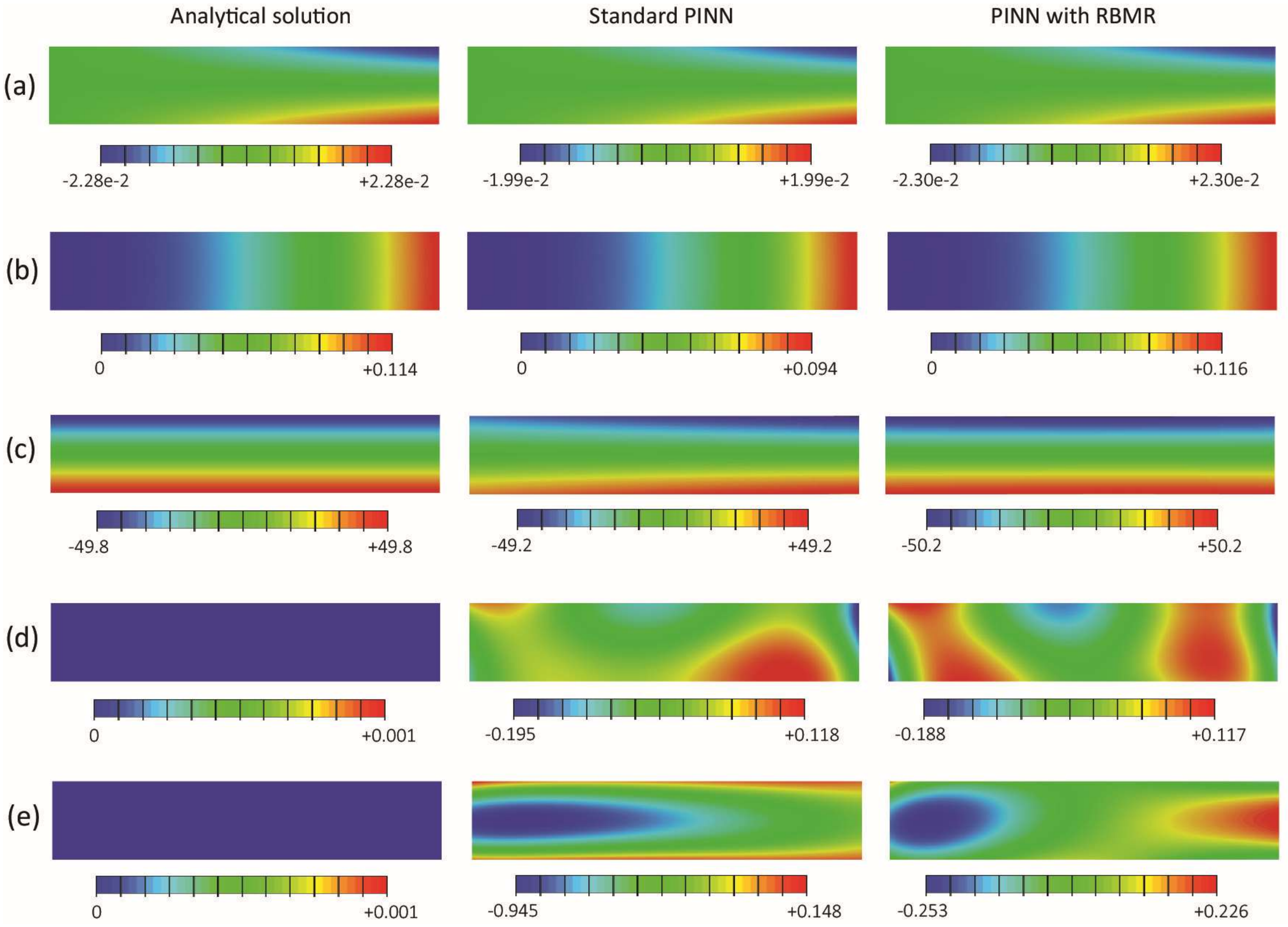

**Fig. 7**. Some snapshots of the simulations of the pure bending beam example by using the standard PINN and the PINN with RBMR methods with neural network FFNN2 and sampling points SP1: (a) displacement $u_x\ (m)$, (b) displacement $u_y\ (m)$, (c) stress $\sigma_{xx}\ (kN/m^2)$, (d) stress $\sigma_{yy}\ (kN/m^2)$, and (e) stress $\sigma_{xy}\ (kN/m^2)$. The first, second, and third columns of plots correspond to the analytical solutions, the results from standard PINN, and the results from PINN with RBMR, respectively.

### *4.2. Curved cantilever beam under end loading*

Deformation of a curved cantilever beam carrying an end loading is considered in this example. The problem domain is a quarter annulus as shown in Fig. 8. Assuming plane strain conditions and unit thickness, the analytical stress field for this problem is given as [56]

$$
\begin{aligned}
\sigma_r &= \frac{P}{A}\left(r - \frac{a^2+b^2}{r} + \frac{a^2b^2}{r^3}\right)\sin(\theta) \\
\sigma_\theta &= \frac{P}{A}\left(3r - \frac{a^2+b^2}{r} - \frac{a^2b^2}{r^3}\right)\sin(\theta) \\
\tau_{xy} &= \frac{P}{A}\left(-r + \frac{a^2+b^2}{r} - \frac{a^2b^2}{r^3}\right)\cos(\theta)
\end{aligned} \tag{30}
$$

where $(r, \theta)$ are the cylindrical coordinates, $P$ is the end loading, $a$ and $b$ are, respectively, the inner and outer radii of the quarter annulus, and

$$A = a^2 - b^2 + (a^2 + b^2)\log\left(\frac{b}{a}\right) \tag{31}$$

The associated analytical displacement field can be calculated as

$$\begin{aligned} u_x &= -\frac{3M}{2Eh^3}xy \\ u_y &= \frac{3M}{4Eh^3}(x^2 + \nu y^2) \end{aligned} \tag{32}$$

in which $E$ and $\nu$ are the modulus of elasticity and Poisson's ratio, respectively.

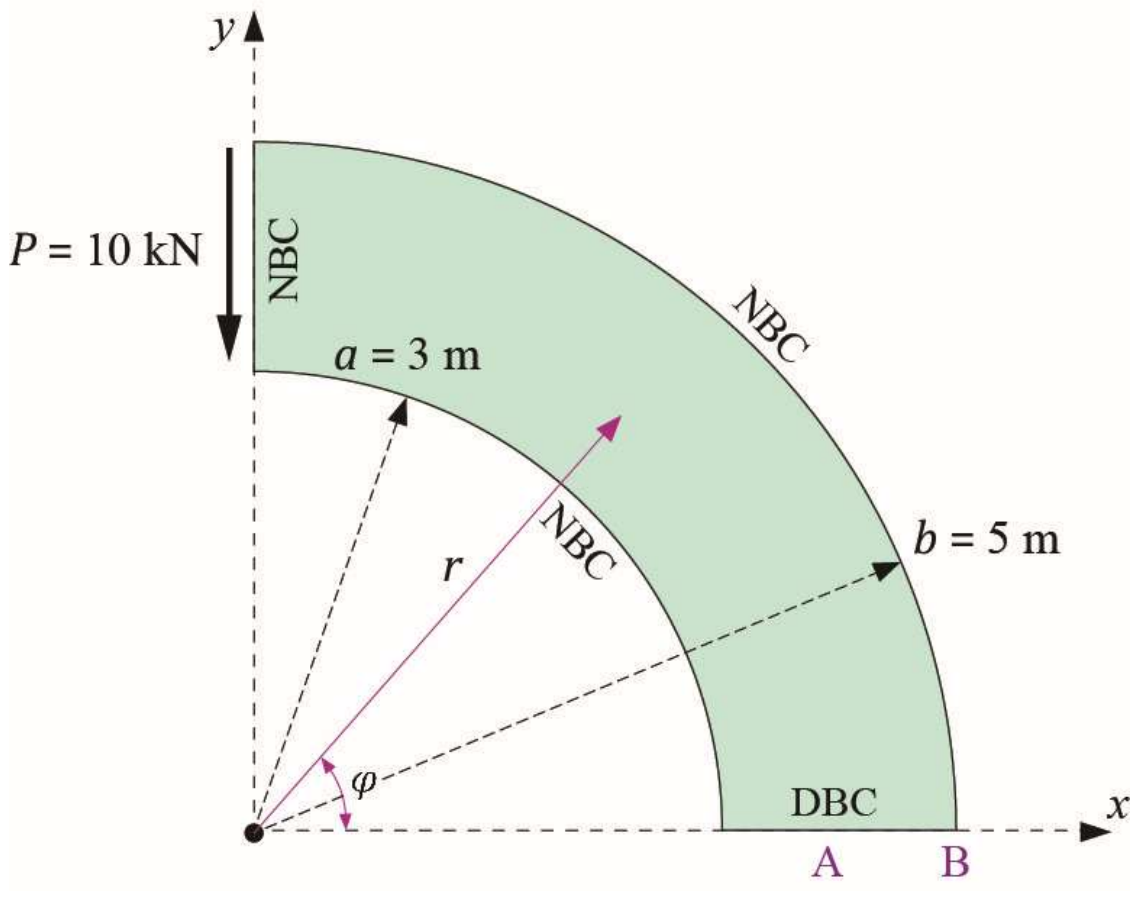

**Fig. 8**. The problem domain in the rotating body example.

As shown in Fig. 8, the numerical simulations impose Dirichlet boundary conditions on the lower edge of the curved beam and Neumann boundary conditions on the remaining boundaries. The boundary conditions are defined based on the analytical displacement and stress values from Eqs. (30) and (32). To implement the RBMR, three known displacement components are chosen as $u_x$ at point $A$, $u_y$ at point $A$, and $u_y$ at point $B$. Two sets of sampling points, shown in Fig. 9, are used in the numerical simulations, including (1) 56 internal sampling and 40 boundary sampling points, denoted SP1, and (2) 616 internal sampling and 120 boundary sampling points, denoted SP2. Also, two FFNNs with three hidden layers are used in different simulations: (1) FFNN1, which contains 2, 8, 32, 8, and 1 neurons in the input, first hidden, second hidden, third hidden, and output layers, respectively, and (2) FFNN2, which contains 2, 16, 64, 16, and 1 neurons in the input, first hidden, second hidden, third hidden, and output layers, respectively. The Adam optimizer is used with an initial learning rate of 0.0001 and an exponential decay with a decay rate of 0.98 and a decay step of 200. The Young's modulus and Poisson's ratio are given as $E = 1000\ \mathrm{kN/m^2}$ and $\nu = 0.3$, respectively.

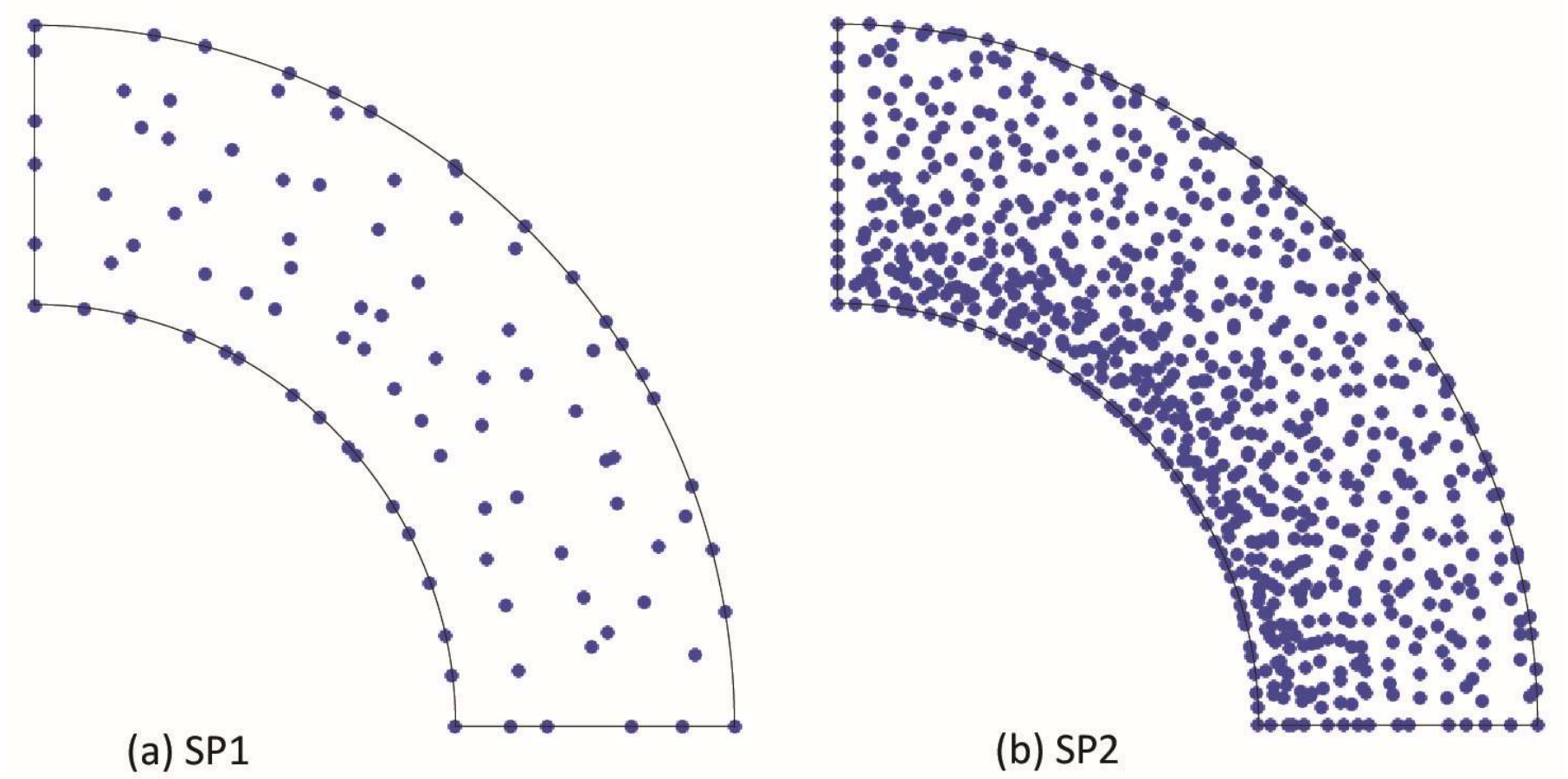

**Fig. 9**. Two sets of sampling points used in the curved beam example: (1) 56 internal sampling and 40 boundary sampling points, denoted SP1, and (2) 616 internal sampling and 120 boundary sampling points, denoted SP2.

Mean error values and their standard deviations from 50 runs of each simulation based on different random initializations are presented in Table 2. The histograms of the displacement and energy error norms for different simulations are also shown in Fig. 10. It can be seen that the proposed method results in smaller mean errors and standard deviations compared to those obtained from the standard PINN.

Table 2. The mean and standard deviation of the displacement and energy error norms for the curved beam example, obtained from the standard PINN and the PINN with RBMR methods

| Neural network | Sampling points | Displacement error norm | | | | Energy error norm | | | |
|---|---|---|---|---|---|---|---|---|---|
| | | standard PINN | | PINN with RBMR | | standard PINN | | PINN with RBMR | |
| | | Mean | Std | Mean | Std | Mean | Std | Mean | Std |
| FFNN1 | SP1 | 0.782 | 0.149 | 0.304 | 0.148 | 0.780 | 0.143 | 0.353 | 0.154 |
| FFNN1 | SP2 | 0.887 | 0.126 | 0.495 | 0.142 | 0.877 | 0.117 | 0.537 | 0.143 |
| FFNN2 | SP1 | 0.583 | 0.146 | 0.120 | 0.045 | 0.588 | 0.143 | 0.155 | 0.049 |
| FFNN2 | SP2 | 0.723 | 0.168 | 0.251 | 0.087 | 0.706 | 0.155 | 0.276 | 0.094 |

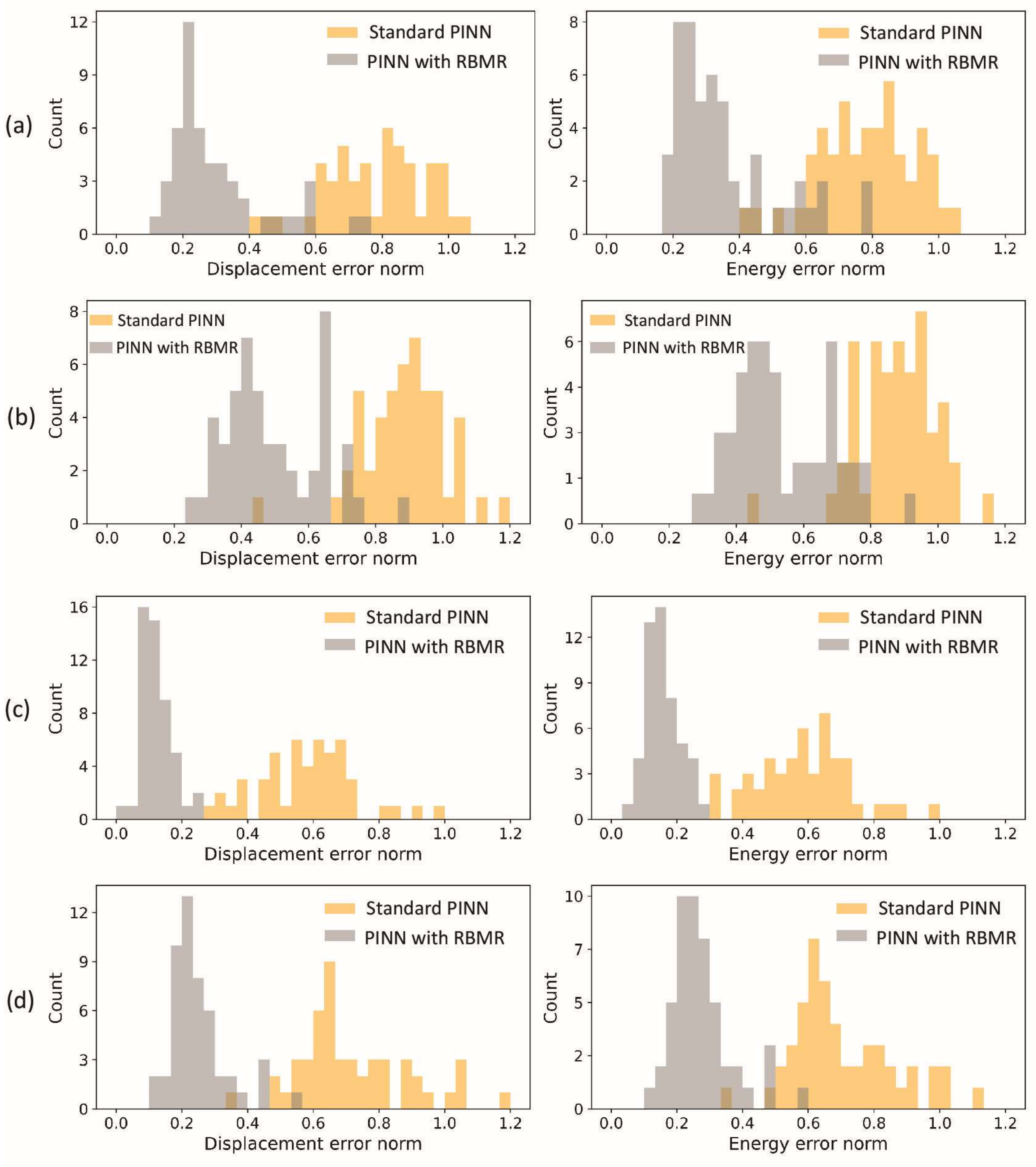

**Fig. 10**. The histograms of the displacement and energy error norms for different simulations of the curved beam example with (a) neural network FFNN1 and sampling points SP1, (b) neural network FFNN1 and sampling points SP2, (c) neural network FFNN2 and sampling points SP1, and (d) neural network FFNN2 and sampling points SP2.

Fig. 11 shows the mean loss curves from the numerical simulations using standard PINN and PINN with RBMR, indicating that both methods show similar convergence behavior.

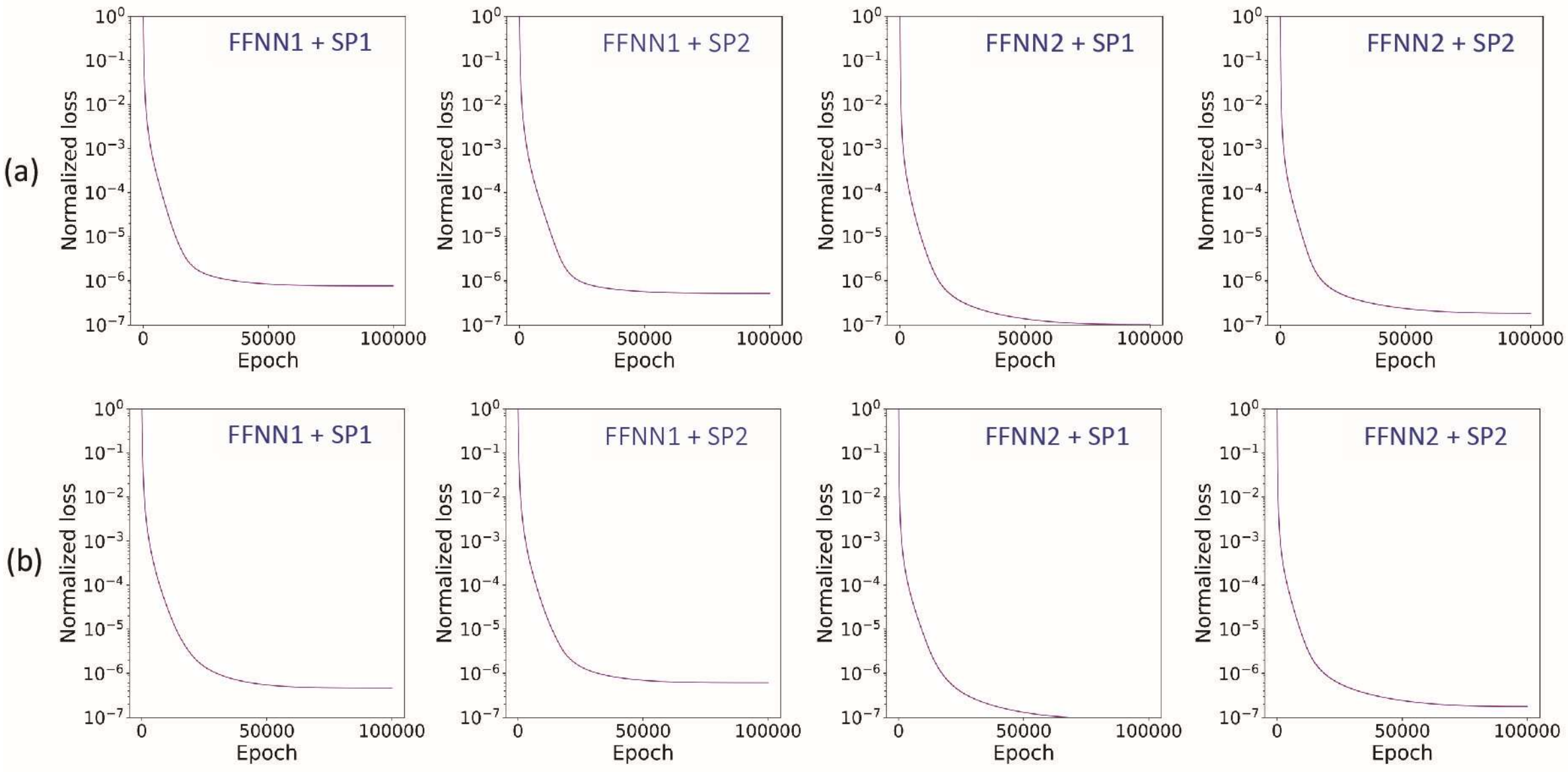

**Fig. 11**. The loss curves for different simulations of the curved beam example with (a) standard PINN and (b) PINN with RBMR.

Fig. 12 shows some snapshots of the simulations with neural network FFNN2 and sampling points SP1 assuming initialization with a seed value of 42. In this case, the obtained displacement norms from the standard PINN and the PINN with RBMR simulations are 0.733 and 0.063 respectively. Similarly, the obtained energy norms from the standard PINN and the PINN with RBMR simulations are 0.716 and 0.100, respectively. As shown in Fig. 12, the results obtained from the PINN with RBMR simulation are in better agreement with the analytical solutions compared to those obtained from the standard PINN simulation.

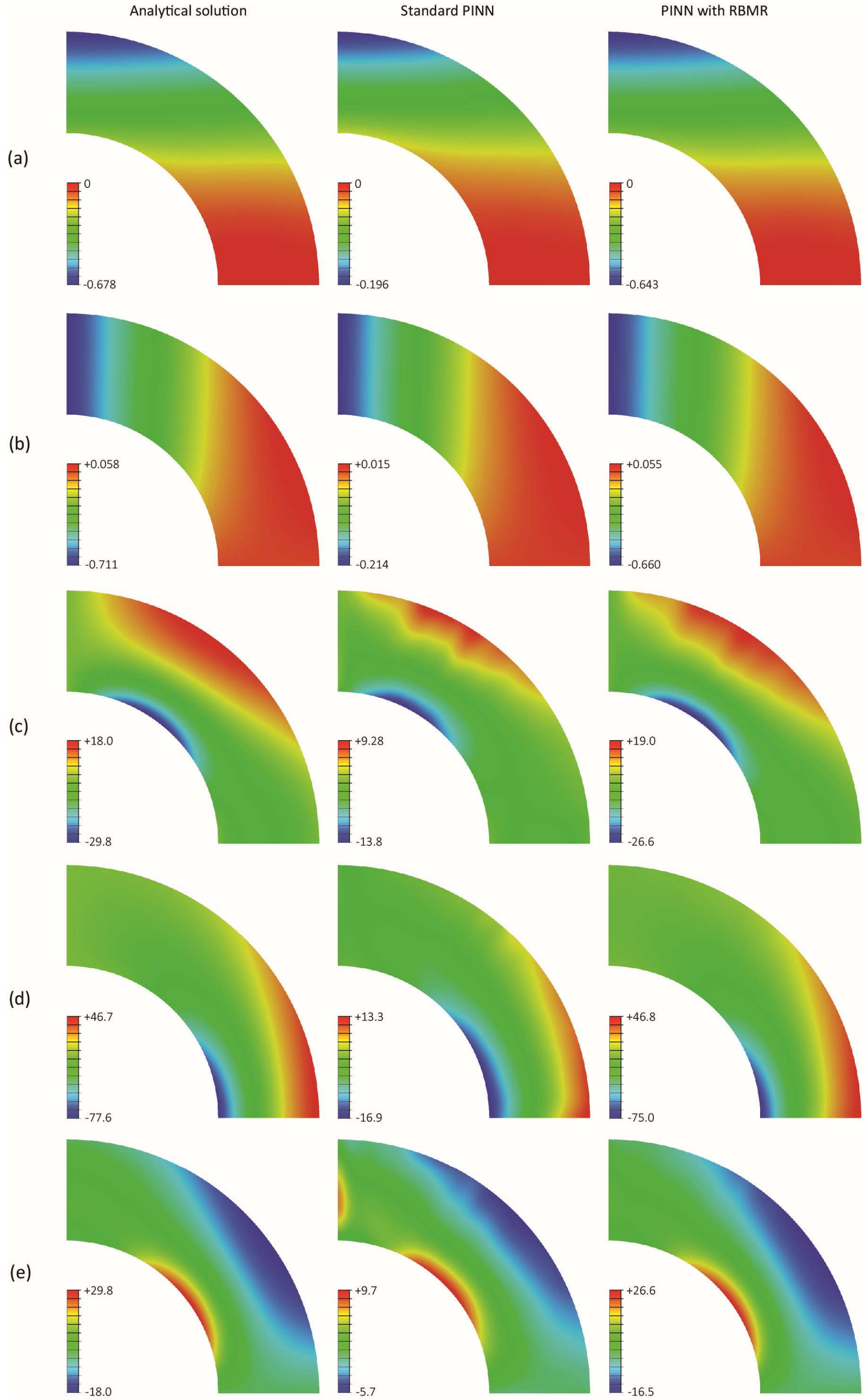

**Fig. 12**. Some snapshots of the simulations of the curved beam example by using the standard PINN and the PINN with RBMR methods with neural network FFNN2 and sampling points SP1: (a) displacement $u_x\ (m)$, (b) displacement $u_y\ (m)$, (c) stress $\sigma_{xx}\ (kN/m^2)$, (d) stress $\sigma_{yy}\ (kN/m^2)$, and (e) stress $\sigma_{xy}\ (kN/m^2)$. The first, second, and third columns of plots correspond to the analytical solutions, the results from standard PINN, and the results from PINN with RBMR, respectively.

## 5. Conclusions

This work proposes a new way of improving the accuracy of physics-informed neural networks for solid mechanics problems by means of an approach termed rigid-body motion reduction (RBMR). PINNs have been known to incorporate physical laws into machine learning models, but they frequently suffer from loss function imbalances and rigid-body motion artifacts. The RBMR approach has remedies to these suffers: it embeds a modified collocation loss function such that a few points are anchored to eliminate the rigid-body motion of the problem analytically. The proposed method is validated against some numerical experiments done on 2D-elastostatic problems involving the case of a pure bending cantilever and a curved cantilever under end loading. The result shows a significant improvement in the performance of accuracy and convergence when compared to the standard PINN approach, implying that incorporating particular domain knowledge into the training process is better.

Although the presentation of the proposed RBMR technique was limited to two-dimensional elasticity problems in this paper, it should be self-evident how to generalize the technique to three-dimensional problems. In this case, six independent displacement components are required to describe the rigid-body motion. The rest of the presented algorithms can be naturally generalized to three-dimensions.